\documentclass[]{interact}

\usepackage{epstopdf}
\usepackage[caption=false]{subfig}

\usepackage[natbibapa,nodoi]{apacite} 
\theoremstyle{plain}

\theoremstyle{definition}

\theoremstyle{remark}

\usepackage{tabularx}
\usepackage{xcolor}
\usepackage{etoolbox} 
\usepackage{graphicx}
\usepackage{hyperref}
\usepackage{multirow}

\newtoggle{anonymous}
\togglefalse{anonymous}

\newcommand{\pvalue}{p\mbox{-value}}

\begin{document}

\title{The optimal placement of the head in the noun phrase. The case of demonstrative, numeral, adjective and noun}

\iftoggle{anonymous}{}
{ 
  \author{
  \name{Ramon Ferrer-i-Cancho\thanks{CONTACT Ramon Ferrer-i-Cancho. Email: rferrericancho@cs.upc.edu}}
  \affil{Quantitative, Mathematical and Computational Linguistics Research Group, Departament de Ci\`encies de la Computaci\'o, Universitat Polit\`ecnica de Catalunya (UPC), Barcelona, Spain. ORCiD: 0000-0002-7820-923X}
  }
}

\maketitle

\begin{abstract}
The word order of a sentence is shaped by multiple principles. The principle of syntactic dependency distance minimization is in conflict with the principle of surprisal minimization (or predictability maximization) in single head syntactic dependency structures: while the former predicts that the head should be placed at the center of the linear arrangement, the latter predicts that the head should be placed at one of the ends (either first or last). A critical question is when surprisal minimization (or predictability maximization) should surpass syntactic dependency distance minimization. In the context of single head structures, it has been predicted that this is more likely to happen when two conditions are met, i.e. (a) fewer words are involved and (b) words are shorter. Here we test the prediction on the noun phrase when it is composed of demonstrative, numeral, adjective and noun. We find that, across preferred orders in languages, the noun tends to be placed at one of the ends, confirming the theoretical prediction. We also show evidence of anti locality effects: syntactic dependency distances in preferred orders are longer than expected by chance.
\end{abstract}


\section{Introduction}

\label{sec:introduction}

One of the biggest challenges of language research is to generate predictions that can be confirmed in spite of the tremendous diversity of world languages \citep{Skirgaard2023a, Evans2009a}. 
An even greater challenge is to generate predictions that hold in a wide range of communication systems of non-human species. Information theory and quantitative linguistics have demonstrated that this challenge may not be as difficult as one may think {\em a priori} \citep{Semple2021a}.

Consider the magnitude of a word type, e.g., its duration or its number of phonemes, in the vocal (spoken languages) or gestural modality (signed languages).
The principle of compression is the pressure to reduce the magnitude of a word type as much as possible \citep{Ferrer2012d}. 
The optimal coding theorem by \citet{Ferrer2019c} predicts that more frequent words should tend to have smaller magnitude. In particular it predicts Zipf's law of abbreviation, namely that more frequent words should tend to be shorter. In detail, that theorem is formulated using two main concepts. First, $L$, the average magnitude of a word type, 
\begin{equation*}
L = \sum_{i}p_i l_i,
\end{equation*}
where $p_i$ is the probability or the relative frequency of the word type $i$ and $l_i$ is its magnitude. Second, $\tau(p_i, l_i)$, the Kendall $\tau$ correlation between the frequency of a word type and its magnitude. A negative correlation, i.e. $\tau(p, l) < 0$, is the operational definition of Zipf's law of abbreviation \citep{Zipf1949a}. The theorem states that
\footnote{One may argue that this prediction is already available in classic information theory \citep{Cover2006a} or in popular publications about the efficiency of languages \citep{Gibson2019a,Levshina2022b}. Classic information theory makes many assumptions on how to code for words in order to achieve optimality that languages do not satisfy strictly (e.g., unique segmentation, non-singularity). Those assumptions yield very precise predictions about the functional dependency between $p_i$ and $l_i$ that only have been verified in a few languages \citep{Torre2019a,Hernandez2019b}.
Ferrer-i-Cancho et al's \citeyearpar{Ferrer2019c} optimal coding theorem gets rid of these assumptions. Furthermore, standard information theory assumes that words consists of discrete units and thus does not address directly the problem, as Ferrer-i-Cancho et al's theorem does, of how to deal with word durations (see \citet{Ferrer2019c} for further details). As for popular publications on language efficiency, notice that they do not add any new mathematical prediction concerning the minimization of $L$ with respect to classic information theory \citep{Cover2006a} or concerning the law of abbreviation; for that reason, that optimal coding theorem is, at present, the most suitable mathematical framework for the large diversity of communication systems in humans and other species. Classic information theory was not developed to understand natural communication systems but to solve engineering problems.} 
\begin{quotation}
If $L$ is minimized, namely, coding is optimal, then $\tau(p_i, l_i) \leq 0$. 
\end{quotation}  
As expected, $\tau(p, l) < 0$ (Zipf's law of abbreviation), has been confirmed without exceptions in samples of languages covering about $13\%$ of world's languages \citep{Bentz2016a, Petrini2022b} and has been confirmed in a wide range of species with interesting exceptions \citep{Semple2021a}. \footnote{The reader can check an updated list of publications on (dis)confirmation of Zipf's law of abbreviation in other species at \href{https://cqllab.upc.edu/biblio/laws/}{Bibliography on laws of language outside human language
}.} 
Here we will show how the principle of compression and this powerful theorem can help one to predict the likely placement of a nominal head in the noun phrase in a large sample of diverse languages. Before, we will review two distinct word order principles that turn out to be in conflict. 

\begin{figure}
\begin{center}
\includegraphics[width = 0.8 \textwidth]{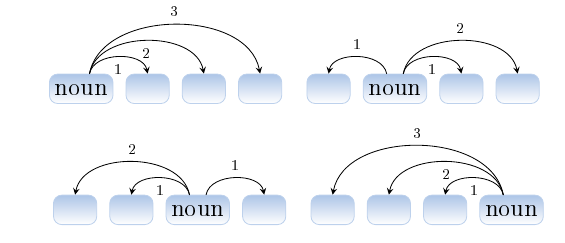}
\end{center}
\caption{\label{fig:noun_phrase_with_3_dependents} All the possible placements of the head for the syntactic dependency structure of a noun phrase formed by a noun and its three dependents. The total sum of dependency distances ($D$) is minimized when the noun is placed at the center forming a balanced structure ($D = 4$) and maximized when it is placed at one of the ends forming a bouquet ($D = 6$). }
\end{figure}

We begin with the currently popular principle of syntactic dependency distance minimization, namely the pressure to reduce the distance between syntactically related words \citep{Ferrer2004b}. When such distance is measured in words (namely consecutive words are at distance one, words separated by an intermediate word are at distance two and so on), various predictions can be made based on the optimal placement of the head. In single head structures on $n$ words, the principle predicts that the root (the only head) should be placed at the center of the linear arrangement. \footnote{Namely in position $\lceil n/2 \rceil$ if $n$ is odd or in positions $n/2$ and $n/2+1$ if $n$ is even, where $n$ is the length of the sequence.}
Figure \ref{fig:noun_phrase_with_3_dependents} shows all possible placements of the head and the corresponding value of $D$, the sum of dependency distances, for a noun phrase consisting of a nominal head and three other dependents. The optimal placement of the noun is at the center (in positions 2 or 3 of the linear arrangement), where the sum of dependency distances is minimized.

In more complex syntactic structures, the addition of the assumption that the linear arrangement has to be planar (namely, no crossing syntactic dependencies), predicts that the root vertex should be placed at the center of the linear arrangement in the sense that half of its subtrees should be arranged before it and the other half after it \citep{Iordanskii1987a, Hochberg2003a, Alemany2021a}.  \footnote{This argument about the optimal placement is applied recursively to roots of the immediate subtrees of the main root until the subtree contains just one vertex.} 
In the case of the order of the subject, object and verb, the root is the verb and the optimal placement of the main verb is in between the subject and the object constituents. As a low number of syntactic dependency crossings (almost planarity) is, to a large extent, a side-effect of syntactic dependency minimization \citep{Gomez2016a,Gomez2019a}, the planarity assumption may not be necessary for the argument to hold in sentences where syntactic dependency distance minimization is acting strongly. 

A less popular word order principle is surprisal minimization or predictability maximization \citep{Ferrer2013f}. In single head structures, the principle predicts that the head should be placed at one of the ends under the assumption that the goal is to reduce the surprisal of either the head or the dependents. If the goal is just to reduce the surprisal (increase the predictability) of the head, then the head should be placed as late as possible. If the goal is to reduce the surprisal (or increase the predictability) of the dependents, the head should be placed as early as possible. If the target (head or dependents) is unknown, that principle predicts that the head should be placed at one of the ends. The particular case of a noun and three dependents is illustrated in Figure \ref{fig:noun_phrase_with_3_dependents}.
In more complex syntactic structures, the minimization of the surprisal of heads predicts left branching (all heads, including the root, are preceded by their dependents) while the minimization of the surprisal of the dependents predicts right branching (namely all dependents are preceded by their head) \citep{Ferrer2014e}. In the case of the order of the subject, object and verb, the root is the verb and two scenarios appear: if the target is the head, the optimal placement of the verb is then after the subject and the object constituents as in the majority of languages or families \citep{Hammarstroem2016a}; if the target are the verbal arguments, the optimal placement of the verb is then before the subject and the object constituents. Put differently, the most attested placement of the verb is at the end of the triple in terms of the dominant order of subject-verb-object \citep{Hammarstroem2016a}, which is expected by surprisal minimization of the verb \citep{Ferrer2013f}.

\subsection{Single-head structures}

A conclusion of the predictions of these two word order principles on single head structures is that they are in conflict. Syntactic dependency distance minimization predicts that the head should be put at the center of the linear arrangement while the minimization of surprisal predicts a placement at one of the ends \citep{Ferrer2013f}. 
The conflict is extreme in single head structures: $D$, the total sum of dependency distances, is maximized when the head is placed at one of the ends according to the principle of surprisal minimization, which is totally against the principle of syntactic dependency minimization (Figure \ref{fig:noun_phrase_with_3_dependents}). \footnote{In single head structures of $n$ words, $D$, the total sum of dependency distances, depends on the position $\pi$ of the head ($1 \leq \pi \leq n$), \citep{Ferrer2013e}
namely 
\begin{equation*}
D(\pi) = \pi^2 - (n+1)\pi + {n+1 \choose 2},
\end{equation*}
where $\pi$ ranges between $1$ and $n$.
In these structures, the placement of the head at one of the ends according to surprisal minimization implies that $D$, the sum of dependency distances, is maximum, i.e. \citep{Ferrer2013b,Ferrer2020a} 
\begin{equation*}
D = D(1) = D(n) = D_{max} = {n \choose 2},
\end{equation*}
contrasting with the minimum dependency distances that are obtained when the head is put at the center according to the principle of syntactic dependency distance minimization, where $D$ is minimized and achieves \citep{Iordanskii1974a,Ferrer2020a}
\begin{equation*}
D = D_{min} = \left\lfloor \frac{n^2}{4} \right\rfloor. 
\end{equation*}
}
When the head of a single head structure is placed at one of the ends, the syntactic dependency structure is called a bouquet, when the head is placed at the center, the syntactic dependency structure is called balanced \citep{Courtin2019a}. See Figure \ref{fig:noun_phrase_with_3_dependents} for all possible configurations in a noun phrase.

Given the conflict above, the following scenarios are possible in single head structures: 
\paragraph*{Scenario 1.} 
Syntactic dependency distance minimization and surprisal minimization are equally strong and then the head can be placed in any position (all positions are equally likely) while syntactic dependency distances are neither shorter nor longer than expected by chance.
\paragraph*{Scenario 2.} 
Syntactic dependency distance minimization is weaker than surprisal minimization and the head tends to be placed at one of the ends while syntactic dependency distances are longer than expected by chance. The syntactic dependency structure should tend to be a bouquet. 
\paragraph*{Scenario 3.} 
Syntactic dependency distance minimization is stronger than surprisal minimization and then the head tends to be placed at the center while syntactic dependency distances are shorter than expected by chance. The syntactic dependency structure should tend to be balanced.\\
~\\

It has been predicted that the likelihood of Scenario 2 is maximized when online memory constraints are less taxing, in particular, when two of the following conditions are met \citep{Ferrer2014a,Ferrer2014e}
\begin{itemize}
\item
$n$ is small ($n \leq 4$). The reason is multiple. First, a small $n$ implies that the memory limitations leading to the principle of dependency minimization, i.e. decay of activation and interference \citep{Liu2017a,Gildea2010a}, are more likely to be less taxing. Second, the sequence fits the size of working memory \citep{Cowan2001a} and is more likely to form a chunk and then be processed as a bundle \citep{Contreras2022a}.   
\item
Words are shorter. The reason is that the true distances (those measured in syllables or phonemes rather than in words as it is customary in quantitative linguistics) are likely to be shorter and thus alleviate the cost of a suboptimal placement of the head according to the principle of syntactic dependency distance minimization \citep{Ferrer2014e}. An example can be found in Romance languages, that are SVO languages (and thus place the head verb optimally according to syntactic dependency distance minimization) but are SOV when O is a clitic (a suboptimal placement of the head according to syntactic dependency distance minimization). Clitics are short words. \footnote{This approach to the placement of clitics relative to the placement of the verb (a bias for arguments to precede the verb under certain conditions) is in contrast with Wackernagel's law, that claims that clitics tend to be placed in second absolute position \citep{Wackernagel1892a, Anderson1993a}.}
\end{itemize} 
\citet{Courtin2019a} investigated syntactic dependency substructures of trigrams in Chinese, English, French and Japanese. They found that Scenario 3 (balanced structures) are not particularly frequent compared to other structures (bouquet and multiple head structures, i.e. zigzag and chain) while Scenario 2 (bouquets) occurs more frequently in actual syntactic dependency structures than in artificially generated ones. \citet{Courtin2019a} admitted to not having a satisfactory explanation for their findings. However, the high frequency of bouquets is consistent with our theoretical prediction of Scenario 2 under the conditions indicated above.  

In another study, \citet{Ferrer2019a} focused on sentences of length $n=3$ and $n=4$ words, that represent a subset of all the trigrams or quadrigrams of a corpus, but they used a much larger sample of languages and linguistic families compared to \citet{Courtin2019a}. Critically, \citet{Ferrer2019a} neglected dependency direction. If dependency direction is disregarded in a rooted tree, one obtains a free tree following the standard terminology of graph theory. Disregarding dependency direction in a single head structure yields a star tree. When $n=3$, the free structure is both a linear and a star tree; when $n=4$, the free structure can be either a star tree or a linear tree \citep{Ferrer2019a}. \citet{Ferrer2019a} found that
\begin{itemize}
\item
Syntactic dependency distances that are neither longer nor shorter than expected by chance, consistently with Scenario 1, were found for most languages.
\item
Syntactic dependency distances longer than expected by chance, consistently with Scenario 2, were found in star trees but never for linear trees. 
\item
Syntactic dependency distances shorter than expected by chance, consistently with Scenario 3, were found more often for linear trees than for star trees. 
\end{itemize}
The conclusions of this study have been confirmed by other studies. First, by means of $\Omega$, an optimality score that tends to give a 0 in case of Scenario 1,   
a negative value in case of Scenario 2 and a positive value in case of Scenario 3 \citep{Ferrer2020b}. A negative $\Omega$ (in the sense of a significantly small $\Omega$) appeared only in sentences of length $3$ or $4$ in certain languages. Second, dependency distances distributed according to a null model (that assumes that all permutations of the sentence are equally likely) have been found in short sentences \citep{Petrini2022c}, consistently with Scenario 1.

In an evolutionary context, syntactic dependency distances in sequences of at most 4 words has been increasing over time in State of the Union Addresses \citep{Liu2022a}, suggesting a tendency towards Scenario 1 or 2 for that genre in English. \footnote{The study does not control for sentence length and thus it is not possible to conclude if dependency distances end up being just the ones expected in a random shuffling of the words of the sentence or greater. However, such a control is easy with the $\Omega$ dependency closeness score, that yields a negative value in case of anti dependency distance effects \citep{Ferrer2020b}.} Both \citet{Ferrer2019a} and \citet{Liu2022a} neglected the identity of the root vertex (whether it is the hub or a leaf), which is a limitation for testing directly the predictions of the theoretical apparatus presented above. 

\begin{figure}
\begin{center}
\includegraphics[width = 0.6 \textwidth]{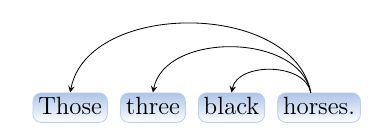}
\end{center}
\caption{\label{fig:noun_phrase_in_English} The syntactic dependency structure of the phrase ``Those three black horses''. ``horses'' is the root and the only head of the phrase. }

\end{figure}

\subsection{The order of demonstrative, numeral, adjective and noun}

Here we borrow the setting of \citet{Dryer2018a} and consider the case of a noun phrase formed by a specific quadruplet: a demonstrative, a numeral, an adjective and a noun, which forms a single head structure. The syntactic dependency structure of an English noun phrase of this sort is shown in Figure \ref{fig:noun_phrase_in_English}. In the language of graph theory, such a structure is a star tree where the root is the hub vertex.
 
In this setting, Dryer proposed the principle of intracategorial harmony: {\em ``The demonstrative, numeral, and adjective tend to all occur on the same side of the noun.''}. This descriptive principle is an instantiation of the explanatory principle of surprisal minimization reviewed above. The principle of surprisal minimization is grounded on information theoretic mathematical properties \citep{Ferrer2013f}. In the context of the order of subject, object and verb, the principle of surprisal minimization predicts a preference for verb last that has been confirmed in acceptability experiments on SOV languages \citep{Ferrer2023a}. 

Notice that this quadruplet setting matches the two conditions that are required to maximize the likelihood of Scenario 2 because $n$ is obviously small ($n=4$) and words are expected to be short. To see the latter, we build first a theoretical argument that we validate later on. The theoretical argument is that the optimal coding theorem by \citet{Ferrer2019c} predicts that more frequent words should tend to be shorter. In our case study, that optimal coding theorem predicts that demonstratives and numerals are short given their high frequency in languages. As for numerals, notice that their length varies according to the magnitude of the number they represent but numbers of low magnitude tend to be the most frequent across languages \citep{Dehaene1992a,Coupland2011a}.
Applying the optimal coding theorem by \citet{Ferrer2019c}, we infer that numbers of high frequency are likely to be shorter. \footnote{Consistently, numbers of higher magnitude happen to be lexicalized, overall, in longer character strings at least in English \citep{Coupland2011a}.} 
To conclude, we predict that, given the characteristics of such a noun phrase, the noun should tend to be placed at one of the ends as in Scenario 2.
Here we will test the prediction based on the frequencies of the $4!=24$ possible preferred orders reported by \citet{Dryer2018a}. 

The remainder of the article is organized as follows. Section \ref{sec:materials} presents the data we borrow from \citet{Dryer2018a}. Section \ref{sec:methods} presents the statistical methods to test the prediction and investigate other aspects of Scenario 2, namely if syntactic dependencies are longer than expected by chance. Section \ref{sec:results} shows that, across languages, the noun is more likely to be placed at one of the ends, confirming the theoretical prediction. That section also shows that syntactic dependency distances are significantly longer, as expected for Scenario 2. 
Section \ref{sec:discussion} discusses the prediction and suggests future research. 
 
\section{Materials}

\label{sec:materials}

We borrow the frequencies of the preferred order of demonstrative, numeral, adjective and noun from Table 2 of \citet[p. 804]{Dryer2018a}. The essential information is reproduced in Table \ref{tab:aboundances} so that the present article is self-contained. 
Word order frequencies are measured by \citet{Dryer2018a} in three units. The first and simplest is in number of languages that prefer the target word order. 
The second is in the number of genera that contain languages preferring the target word order. The motivation of this unit is that counts of number of languages can be distorted by large genealogical groups. However, this second unit can be distorted by languages in contact with each other. The third is adjusted number of languages, that controls for genealogy and geography (see \citet{Dryer2018a} for further details). \footnote{That measurement is called adjusted frequency by \citet{Dryer2018a} but we borrow the name from \citet{Martin2019a} because it makes explicit the unit of measurement. 
}

\begin{table}
  \caption{\label{tab:aboundances} The frequency of each of the $4! = 24$ possible orderings of demonstrative (D), numeral (N), adjective (A) and noun (n) in Dryer's \citeyearpar{Dryer2018a} dataset. $F$, the total frequency, is shown at the bottom.}  
  \begin{center}
  \begin{tabular}{llll}
  Order & Languages & Genera & Adjusted languages \\
  \hline
  nAND & 182 & 85 & 44.17 \\
DNAn & 113 & 57 & 35.56 \\
DnAN & 53 & 40 & 29.95 \\
DNnA & 40 & 32 & 22.12 \\
nADN & 36 & 19 & 14.8 \\
NnAD & 67 & 27 & 14.54 \\
DnNA & 12 & 10 & 9.75 \\
nDAN & 13 & 11 & 9 \\
nNAD & 11 & 9 & 9 \\
nDNA & 8 & 6 & 5.67 \\
DAnN & 12 & 7 & 5.34 \\
NAnD & 8 & 5 & 4 \\
AnND & 5 & 3 & 3 \\
NnDA & 5 & 3 & 3 \\
AnDN & 5 & 3 & 2.5 \\
DANn & 3 & 2 & 2 \\
NDAn & 2 & 2 & 2 \\
nNDA & 1 & 1 & 1 \\
NADn & 0 & 0 & 0 \\
NDnA & 0 & 0 & 0 \\
ADnN & 0 & 0 & 0 \\
ADNn & 0 & 0 & 0 \\
ANDn & 0 & 0 & 0 \\
ANnD & 0 & 0 & 0 \\

  $F$ & 576 & 322 & 217.4 \\
  \end{tabular}
  \end{center}
\end{table}

\section{Methods}

\label{sec:methods}

We consider the general setting of a headed phrase that is a continuous constituent and is formed by a sequence of $n$ words.
We will apply this setting to the particular case of a noun phrase formed by demonstrative, numeral, adjective and noun ($n=4$). 

\subsection{The null hypothesis}

Our null hypothesis is that the selection of a preferred order is made by choosing one of the possible orderings of the phrase uniformly at random and independently of the outcome of other selections. By uniformly at random we mean that all the $n!$ linear orderings are equally likely. 
We will refer informally to this null hypothesis as random shuffling. 

\subsection{Statistical tests}

The following tests are applied to frequencies of a certain word order. Frequencies can be either frequencies of a word order in a corpus of a certain language or they can be frequencies of word order among preferred or dominant word orders over languages. In the former, we are dealing with frequencies that reflect word order variation within a language. In the latter, we are dealing with variation of preferred order across languages. In this article, we deal with the latter on world-scale preferred orders for the order of demonstrative, numeral, adjective and noun. 
For that reason, we also use the generic term number of instances, to refer to one of these notions of frequency.
Given a certain definition of frequency, our implicit working hypothesis is that word order variation reflects primarily cognitive preferences in addition to other factors such as evolutionary paths or random fluctuations or manifestations of path-dependent stochastic processes \citep{Arthur1994}.

\subsubsection{A test for the preference for head initial or head final orders}

\label{subsec:preference_for_head_initial_or_final}

The principle of predictability maximization (surprisal minimization) predicts that the head should be placed at one of the ends of the sequence \citep{Ferrer2013f}. In our application case, the noun has to be placed at one of the ends of the noun phrase.  

There are $2(n-1)!$ orders of the sequence where the head is placed at one of the ends of the sequence, i.e. first (in position $1$) or last (position $n$). Then the probability that a uniformly random order has the head at one of the ends is
\begin{equation*}   
p_{1,n} = \frac{2(n-1)!}{n!} = \frac{2}{n}.
\end{equation*}
When $n = 3$, $p_{1,3}=2/3$.
In our application to a nominal phrase, $n=4$ and $p_{1,4} = 1/2$.

We define $f_i$ as the frequency of a word order, where $i = 1, 2,..., n!$. The total frequency is  
\begin{equation*}
F = \sum_{i=1}^{n!} f_i.
\end{equation*}
In our application, $f_i$ is the frequency of a preferred word order measured in one of the units employed by \citet{Dryer2018a} (recall Section \ref{sec:materials}).
We define $g_{1,n}$ as the frequency of orders where the head has been placed at one of the ends (first or last). Frequencies are integer when measured in languages or families and can be non-integer when measured with adjusted number of languages (Table \ref{tab:aboundances}).
 
First, we present a statistical test to check if $g_{1,n}$ is significantly large when $g_{1,n}$ is integer. A significantly large value would confirm the predictions of the principle of predictability maximization (surprisal minimization). Under the null hypothesis above, $g_{1,n}$ follows a binomial distribution with parameters $F$ and $p_{1,n}$. Accordingly, here we apply a right-sided binomial test to check if $g_{1,4}$ is significantly large.

Second, we explain how to adapt the previous test to the case that $g_{1,n}$ is not integer because that test expects that $g_{1,n}$ is integer. $g_{1,4}$ can be non-integer when the frequency of preferred word orders is measured with Dryer's adjusted number of languages. Our solution is to apply four binomial tests resulting from all the transformations of $F$ and $g_{1,4}$ into integer numbers. 
A non-integer number $x$ can be transformed into $\lfloor x \rfloor$, its nearest integer below (e.g., 10.3 becomes $\lfloor 10.3 \rfloor = 10$) and also into $\lceil x \rceil$, its nearest integer above (e.g., 10.3 becomes $\lceil 10.3 \rceil = 11$). The computation of $\lfloor x \rfloor$ is known as flooring while the computation of $\lceil x \rceil$ is known as ceiling. Then we apply four simple binomial tests where $F$ and $g_{1,4}$ are replaced in each by  
\begin{enumerate}
\item
  $\lfloor F \rfloor$, $\lfloor g_{1,4} \rfloor$.
\item
  $\lfloor F \rfloor$, $\lceil g_{1,4} \rceil$.  
\item
  $\lceil F \rceil$, $\lfloor g_{1,4} \rfloor$.  
\item
  $\lceil F \rceil$, $\lceil g_{1,4} \rceil$.    
\end{enumerate}
If each of the four tests yields significance, one can safely conclude that $g_{1,4}$ is significantly large.

\subsubsection{A test for anti syntactic dependency distance minimization effects revisited}
\label{subsec:anti_locality_test}

The binomial test above is equivalent to another test with another objective but same implementation. The aim of that other test was to detect evidence that syntactic dependency distances are larger or longer than expected by chance, against the principle of syntactic dependency distance minimization \citep{Ferrer2019a}. When used to test if syntactic dependency distances are longer than expected by chance, this test is based on a statistic $f_+$ that is the number of times $D$ is larger than $D_r$, the expected value of $D$ in random shufflings of the words of the sentence. When used to test if syntactic dependency distances are shorter than expected by chance, the test is based on a statistic $f_-$ that is the number of times $D$, the total sum of dependency distances, is smaller than $D_r$. 
\iftoggle{anonymous}{}{
These binomial tests were introduced first by \citet{Ferrer2019a}. \footnote{\citet{Futrell2020a} presented an equivalent ``sign'' test but they did not acknowledge the precedent of Ferrer-i-Cancho and Gomez-Rodriguez, who released publicly their work in July 2019 \url{https://arxiv.org/abs/1906.05765v1}, had it published online in the Journal of Quantitative Linguistics in 22 August 2019 although their article did not get an issue until 2021.}
}

When $n=4$ and the free tree structure is a star tree, $f_+$ and $f_-$ are binomially distributed under the null hypothesis. Specifically, $f_+$ and $f_-$ follow a binomial distribution with parameters $N$ and $1/2$, being $N$ the number of instances that have a star tree structure. 

Now let us come back to the binomial test for the preference for head initial or head final orders (Section \ref{subsec:preference_for_head_initial_or_final}). It is easy to see that, when the $f_i$'s are integer, $F=N$, $g_{1,4}=f_+$ and $p_{1, n}=1/2$. Therefore, the binomial test for a preference for head initial or head final orders is equivalent to a test for anti syntactic dependency distance minimization effects if the units of frequency are the same when $n=3$ or $n=4$. If the $f_i$'s are non-integer, the equivalence between both tests is an approximation.  
In plain words, if we find evidence that the head tends to be placed at the ends of the linear arrangement we are also finding that syntactic dependency distances are longer than the random baseline.

\subsection{Confidence intervals for proportions.}

Although the binomial test is the most accurate way to determine if $g_{1,4}$ is significantly large (at least for frequencies in languages or genera), we wish to calculate confidence intervals to visualize the separation between the proportions $g_{1,4}/F$ and $1 - g_{1,4}/F$. 

Suppose a binomial distribution with parameters $F$ and $\pi$. Let $P(x)$ be the cumulative distribution function of a random variable $x$ and $Q(x) = P^{-1}(x)$ the corresponding quantile function. With a significance level $\alpha$, the confidence interval is $Q(\alpha/2, 1 - \alpha/2)/F$. The confidence interval for a certain proportion is obtained by setting $\pi$ to that proportion. \footnote{ 
This method is available in the {\tt epitools} R package. 
}
Here we use $95\%$ confidence intervals (i.e. $\alpha = 0.05$). For the particular case of adjusted number of languages, that is non-integer, we simply round $F$.

\subsection{$\left<D\right>$, the mean syntactic dependency distance}

Let us assume a star tree structure with $n \in \{3,4\}$. The binomial test on $g_{1,n}$ (and equivalently on $f_+$) aims to find evidence that heads tend to be placed at one of the ends or that syntactic dependency distances are longer than expected by chance but does not pay attention to the average length of the dependencies, that is defined as 
\begin{equation*}
\left<D\right> = \frac{1}{F} \sum_{i=1}^F D_i, 
\end{equation*}
where $D_i$ is the sum of dependency distances of the $i$-th instance. However, $g_{1,n}$ (and equivalently on $f_+$) and $\left<D\right>$ are very closely related. Hereafter we use $g_{1,n}$ (which can be replaced equivalently by $f_+$). 
When $n=3$, $D_i \in \{2,3\}$.
Then 
\begin{eqnarray}
\left<D\right> & = & \frac{1}{F} \left[ 2(F - g_{1,n}) + 3g_{1,n} \right] \nonumber \\
               & = & 2 - \frac{g_{1,n}}{F}. \label{eq:average_sum_of_syntactic_dependency_distances_3_vertices}
\end{eqnarray}
When $n=4$, $D_i \in \{4,6\}$.
Then 
\begin{eqnarray}
\left<D\right> & = & \frac{1}{F} \left[ 4(F - g_{1,n}) + 6g_{1,n} \right] \nonumber \\
               & = & 2\left[2 - \frac{g_{1,n}}{F} \right]. \label{eq:average_sum_of_syntactic_dependency_distances_4_vertices}
\end{eqnarray}
These findings have another important consequence: testing if $\left<D\right>$ is significantly large is equivalent to testing if $g_{1,n}$ (or equivalently $f_+$) is significantly large. 

\subsection{The distribution of $\left<D\right>$ under the null hypothesis}

Under the null hypothesis, the expected value of 
$\left<D\right>$ is \citep{Ferrer2018a}
\begin{equation*}
\mu(\left<D\right>) = \mu(D) = D_r = \frac{n^2 - 1}{3}.
\end{equation*} 
Then $\mu(\left<D\right>) = 8/3$ when $n=3$ and $\mu(\left<D\right>) = 5$ when $n=4$.

The standard deviation of $\left<D\right>$ under the null hypothesis, $\sigma(\left<D\right>)$, can be obtained in two ways: applying existing general theory or tailoring a specific solution for Equations \ref{eq:average_sum_of_syntactic_dependency_distances_3_vertices} and \ref{eq:average_sum_of_syntactic_dependency_distances_4_vertices}. Let us begin with the former to illustrate the power and utility of word order theory \citep{Ferrer2018a, Ferrer2020a}. 
As $\left<D\right>$ is an average, 
\begin{equation*}
\sigma(\left<D\right>) = \frac{\sigma(D)}{\sqrt{F}}.
\end{equation*}
According to word order theory, the variance of $D$ under the null hypothesis is \citep{Ferrer2018a} 
\begin{equation*}
V(D) = \frac{n+1}{45}\left[(n-1)^2 + \left(\frac{n}{4}-1\right)n \left<k^2\right>\right]
\end{equation*}
for an arbitrary tree that has $n$ vertices and a second moment of degree about zero that is $\left<k^2\right>$.
If the tree is a star, then $\left<k^2\right> = n - 1$ and then \citep{Ferrer2020a}
\begin{equation*}
V(D) = \frac{1}{180}(n-2)(n-1)(n+1)(n+2).
\end{equation*}
When $n=3$, $V(D) = 2/9$, which yields 
\begin{equation}
\sigma(\left<D\right>) = \frac{2}{9\sqrt{F}}. \label{eq:sigma_under_null_hypothesis_3_vertices}
\end{equation}
When $n=4$, $V(D) = 1$, which yields
\begin{equation}
\sigma(\left<D\right>) = \frac{1}{\sqrt{F}}. \label{eq:sigma_under_null_hypothesis_4_vertices}
\end{equation}
As for the second method, the application of well-known properties of variance\footnote{Given a random variable $X$, its variance satisfies $V[aX+b] = a^2 V[X]$ for any two constants $a$ and $b$ \citep[p. 195]{DeGroot1989a}. } to the definition of $\left<D\right>$ in Equations
\ref{eq:average_sum_of_syntactic_dependency_distances_3_vertices} and \ref{eq:average_sum_of_syntactic_dependency_distances_4_vertices} give
\begin{eqnarray*}
V(\left<D\right>) & = & V \left(2 - \frac{g_{1,n}}{F} \right) \\
                  & = & \frac{1}{F^2} V(g_{1,n})
\end{eqnarray*} 
for $n=3$ and
\begin{eqnarray*}
V(\left<D\right>) & = & V \left(2\left[2 - \frac{g_{1,n}}{F} \right] \right) \\
                  & = & \frac{4}{F^2} V(g_{1,n})
\end{eqnarray*}
for $n=4$. 
As $g_{1,n}$ is binomially distributed with parameters $F$ and $p_{1,n}$ for $n \in \{3, 4\}$, then $V(g_{1,n}) = Fp_{1,n}(1-p_{1,n})$.
What follows confirms the results with the first method: when $n = 3$, $p_{1,3} = 2/3$ and then $\sigma(\left<D\right>) = 2 /(9\sqrt{F})$; when $n = 4$, $p_{1,4} = 1/2$ and then $\sigma(\left<D\right>) = 1 /\sqrt{F}$.   

We define the distance in $\sigma$'s between $\left<D\right>$ and $\mu(\left<D\right>)$ as 
\begin{equation*}
k = \frac{|\left<D\right> - \mu(\left<D\right>)|}{\sigma(\left<D\right>)},
\end{equation*}
which is
\begin{equation*}
k = \frac{9\sqrt{F}}{2} |\left<D\right> - 8/3|
\end{equation*}
for $n=3$ and
\begin{equation*}
k = \sqrt{F} |\left<D\right> - 5|
\end{equation*}
for $n=4$.
According to the $3\sigma$ rule, if $k \geq 3$ then $\left<D\right>$ is significantly larger than $\mu(\left<D\right>)=D_r$, i.e. $\pvalue \leq \alpha = 0.05$.  
To be able to apply the rule safely as a sufficient condition for significance, we must check that our application satisfies the assumptions of the Vysonchanskii-Petunin inequality, from which the rule emerges \citep{Pukelsheim1994a}. The two assumptions are the distribution of $\left<D\right>$ has finite variance (of course, as the possible $D$'s are finite; recall also Equations \ref{eq:sigma_under_null_hypothesis_3_vertices} and \ref{eq:sigma_under_null_hypothesis_4_vertices})
and is unimodal (this is easy to check numerically). 
Notice however that the rule yields a conservative statistical significance test (i.e. it lacks statistical power) compared to the binomial test above, as the $3\sigma$ rule is based on a generic argument that does not take into account the exact distribution under the null hypothesis. 

As an orientation for discussion, we assume a significance level of $\alpha = 0.05$ throughout this article.

\section{Results}

\label{sec:results}

\subsection{The preference for noun first or last}

\begin{figure}
  \begin{center}
  \includegraphics[width = 0.9 \textwidth]{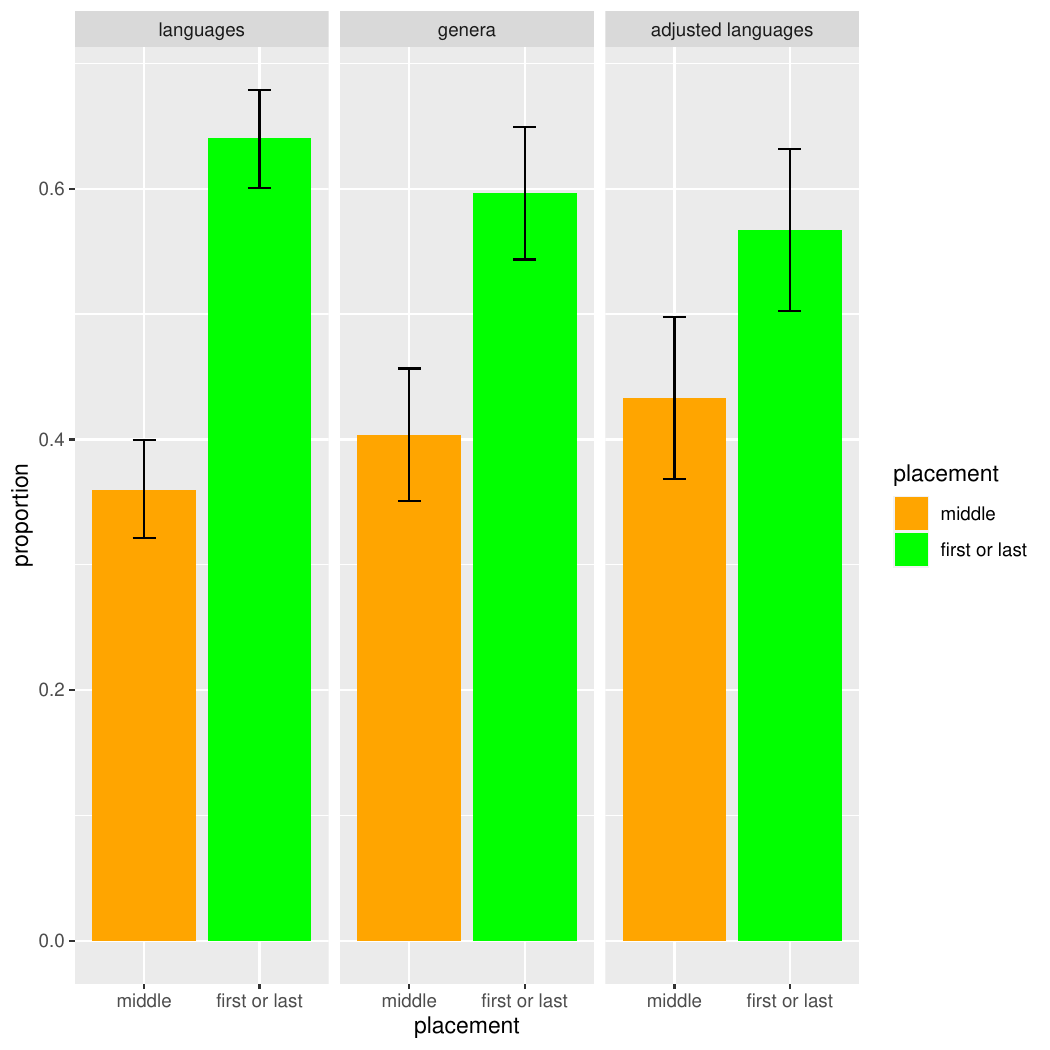}
  \end{center}
  \caption{\label{fig:bar_plots_of_proportion} $g_{1,n}/F$, the proportion of instances where the noun is placed first or last versus, $1 - g_{1,n}/F$, the proportion of instances where the noun is placed in the middle, for each unit of measurement. Error bars indicate the limits of a $95\%$ confidence interval. }
\end{figure}

Figure \ref{fig:bar_plots_of_proportion} shows that placing the noun at one of the ends of the linear arrangement is preferred over placing it in the middle. However, the gap between the two alternative placements of the head decreases as more statistically robust frequency units are used: the gap with languages is greater than that of genera, whose gap is in turn greater than that of adjusted number of languages.

Table \ref{tab:binomial_test} shows that the frequency of preferred orders that place the noun at one of the ends is significantly large independently of the unit of measurement and, in the case of adjusted number of languages, independently of the transformation into integer numbers that is required to apply the binomial test (Table \ref{tab:binomial_test} indicates that $g_{1,n}$ is always significantly large independently of the unit of measurement according to the binomial test). Thus, over preferred orders in languages
\begin{enumerate}
\item
The noun tends to be placed at one of the ends as predicted theoretically. 
\item
The sum of syntactic dependency distances tends to be greater than the one expected under the null hypothesis, i.e. $D>D_r$ tends to be satisfied. 
\item
$\left<D\right>$, the average sum of dependency distances, is larger than expected by the null hypothesis.   
\end{enumerate} 
Therefore, our findings support Scenario 2. 

\begin{table}
\caption{\label{tab:binomial_test} 
The outcome of the right-sided binomial test for $g_{1,4}$, the frequency of noun first or last orders of demonstrative, numeral, adjective and noun, as well as related parameters. The parameters are the unit of measurement of frequency, $g_{1,4}/F$, the relative frequency of noun first or noun last placement and one of the parameters of the binomial distribution, $F$, the total frequency (the other parameter of the binomial distribution is the probability of placing the noun first or last in a random shuffling, that is always $1/2$). When the unit of measurement is adjusted number of languages, $F=217.4$ and $g_{1,4}=123.2$. As the binomial test requires integer numbers, we apply the binomial test to each of the integer transformations that result from flooring or ceiling operations. }
\begin{center}
\begin{tabular}{lllll}
Unit of measurement & $g_{1,4}/F$ & $F$ & $g_{1,4}$ & $\pvalue$ \\
\hline
languages & 0.641 & 576 & 369 & $7.3\cdot 10^{-12}$ \\
genera & 0.596 & 322 & 192 & $3.3\cdot 10^{-4}$ \\
adjusted languages & 0.567 & 217 & 123 & 0.029 \\
adjusted languages & 0.564 & 218 & 123 & 0.034 \\
adjusted languages & 0.571 & 217 & 124 & 0.021 \\
adjusted languages & 0.569 & 218 & 124 & 0.025 \\
\end{tabular}
\end{center}
\end{table}

\subsection{The average sum of dependency distances}

\begin{figure}
  \begin{center}
  \includegraphics[width = 0.9 \textwidth]{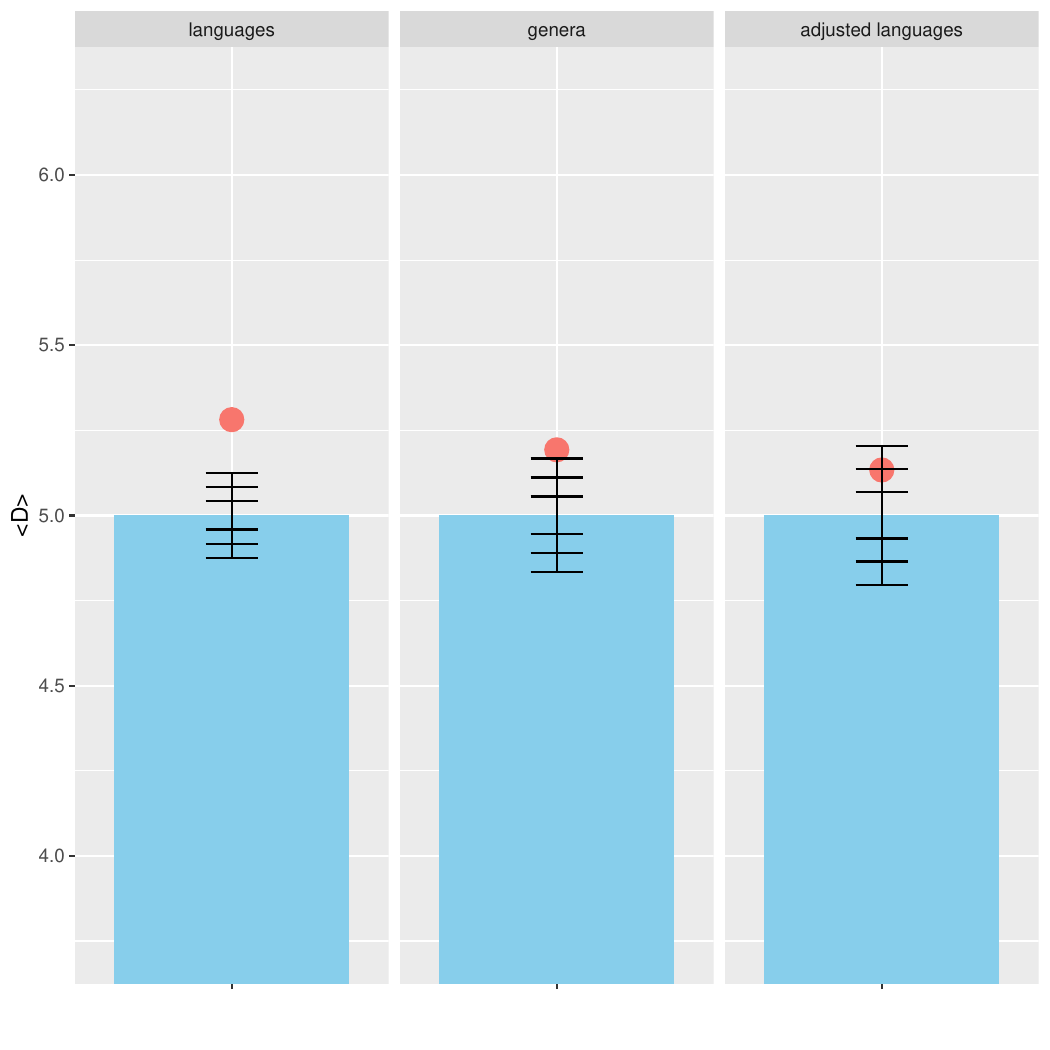}
   \end{center}
  \caption{\label{fig:bar_plots_of_average_D} $\left<D\right>$, the actual average sum of dependency distances (red point) over all instances against the distribution of $\left<D\right>$ in a random shuffling of the words forming the quadruplet (wide blue bars): for each unit of measurement, the height of the bar indicates the expected value $\mu(\left<D\right>) = 5$ and the error bars indicate $\pm k\sigma(\left<D\right>)$ with $k \in \{1,2,3\}$. }  
\end{figure}

Although the binomial test on $g_{1,n}$ suffices to conclude that $\left<D\right>$ is significantly larger than $\mu(\left<D\right>) = 5$ (Section \ref{subsec:anti_locality_test}), we wish to investigate further $\left<D\right>$ and $\mu(\left<D\right>)$ to achieve a visual understanding of the conclusions obtained with the binomial test. 
Figure \ref{fig:bar_plots_of_average_D} shows that, $\left<D\right>$, the actual average sum of syntactic dependency distances is always greater than $\mu(\left<D\right>) = 5$, its expected value in a random shuffling of the words of the quadruplet. As it happens to the gap between $g_{1,n}/F$ and $f_-/F = 1 - g_{1,n}/F$, the gap between $\left<D\right>$ and $\mu(\left<D\right>) = 5$ reduces as more robust units are used: the gap with languages is greater than that of genera, whose gap is in turn greater than that of adjusted number of languages, as expected by the direct association between placement of the head and $D$ (Equation \ref{eq:average_sum_of_syntactic_dependency_distances_4_vertices}). 
Table \ref{tab:average_D} shows the parameters of the distribution of $\left<D\right>$ under the null hypothesis as well as the actual $\left<D\right>$. 
Although the separation between $\left<D\right>$ and $\mu(\left<D\right>)$ is ``small'' (e.g., $\left<D\right> = 5.281$ versus $\mu(\left<D\right>) = 5$), the separation between 
$\left<D\right>$ and $\mu(\left<D\right>)$ is large for counts in languages and genera in the sense they are separated by at least three standard deviations ($k\geq 3$), which warrants the statistical significance according to the $3\sigma$ rule \citep{Pukelsheim1994a}. For adjusted number of languages, $k \approx 2$ and the general $\sigma$-rule is not able to find a significant difference. However, we have already demonstrated with the binomial test on $g_{1,n}$ above, that the difference is significant.

\begin{table}
\caption{\label{tab:average_D}
The average sum of dependency distances, $\left<D\right>)$, and further information about its distribution: its theoretical minimum and maximum ($D_{min}=4$ and $D_{max}=6$, respectively), its expectation and standard deviation under the null hypothesis ($\mu(\left<D\right>) = D_r = 5$ and $\sigma(\left<D\right>) = 1/\sqrt{F}$) and $k$, the separation between $\left<D\right>$ and $\mu(\left<D\right>)$ in units of $\sigma(\left<D\right>)$. }
\begin{center}
\begin{tabular}{llllllll}
Unit of measurement & $F$ & $D_{min}$ & $\mu(\left<D\right>)$ & $\sigma(\left<D\right>)$ & $\left<D\right>$ & $D_{max}$ & $k$ \\
\hline
languages & 576 & 4 & 5 & 0.042 & 5.281 & 6 & 6.75 \\
genera & 322 & 4 & 5 & 0.056 & 5.193 & 6 & 3.46 \\
adjusted languages & 217.4 & 4 & 5 & 0.068 & 5.133 & 6 & 1.97 \\
adjusted languages & 217 & 4 & 5 & 0.068 & 5.134 & 6 & 1.97 \\
adjusted languages & 218 & 4 & 5 & 0.068 & 5.128 & 6 & 1.9 \\
adjusted languages & 217 & 4 & 5 & 0.068 & 5.143 & 6 & 2.1 \\
adjusted languages & 218 & 4 & 5 & 0.068 & 5.138 & 6 & 2.03 \\

\end{tabular}
\end{center}
\end{table}

\section{Discussion}

\label{sec:discussion}

We have confirmed the prediction that the noun should tend to be preceded or followed by all its dependents (a demonstrative, a numeral and an adjective). 
The prediction is based on the principle of compression and the principle of surprisal minimization. The latter is a general version of Dryer's principle of intracategorial harmony \citep{Dryer2018a}. 
We have also demonstrated that this tendency is also a tendency against locality, namely a tendency of syntactic dependencies to be longer than expected by chance, when distances are measured on world-wide word order preferences. 

In Dryer's previous work, the principle of intracategorial harmony is one of five descriptive principles that he used to predict the frequency of the possible orderings of the noun phrase: the more principles are satisfied by an order, the higher the frequency. Here our aim was to predict the more likely placement of the noun applying a theoretical apparatus that has been applied successfully in other domains \citep{Ferrer2019a,Ferrer2020b,Ferrer2014a}.  

Evidence that the effects of the dependency distance minimization principle can also be observed in short spans within noun phrases for Romance languages, has led to cast doubts on the grounding of these effects in memory limitations \citep{Gulordava2015a}, that is a central assumption of our theoretical apparatus (Section \ref{sec:introduction}). However, a pressuring question of such a critical view is: why is that languages show a bias against placing the noun at the center of the noun phrase if that principle pervades the organization of language at all scales? An empirical answer is that Romance languages have been shown to exhibit a high degree of optimality of dependency distances \citep{Ferrer2020b} and thus a more typologically diverse ensemble of languages is required as in our study. A theoretical answer is that the setting of \citet{Gulordava2015a}, i.e. complex noun phrases, does not match the conditions required to observe the anti dependency distance minimization effects as predicted by our theoretical framework (Section \ref{sec:introduction}).   

A bias for noun first or last could also be predicted by a bias favoring noun phrase word orders that transparently reflect constituency structure, or semantic scope relations \citep{Martin2019a,Culbertson2014a,Culbertson2020a}. That bias makes a precise prediction of the preferred order of the whole noun phrase. However, our theoretical apparatus has a more general scope as it can be applied to predict the placement of the head of any phrase that satisfies the conditions presented in Section \ref{sec:introduction} and for that reason it has already been applied to short sentences that do not need to be noun phrases \citep{Ferrer2019a,Ferrer2020b}.

A limitation of the present research article is that the prediction of an initial or final placement of the noun relies on two conditions: (a) the noun phrase is short and that (b) its words are short (Section \ref{sec:introduction}). It is evident that the noun phrases we have examined are short as they consist of just four words. However, we have no direct evidence of condition (b) as we have neither measured word length in that setting nor involved actual word lengths measurements in the argument; we just expect that condition (b) is likely to hold based on partial arguments on the expected short length of common demonstratives and numerals (leaving aside the issue of the length of adjectives). Therefore, we cannot exclude that the actual reason why the theoretical prediction holds is mainly (a). Notice, however, that involving detailed word length information for the 576 languages in Dryer's sample \citep{Dryer2018a} is likely to be a costly endeavor. Thinking in terms of the principle of compression and its predictions yields a shortcut and a preliminary confirmation of the theoretical predictions.

\subsection{Generalizability}

\label{subsec:generalization}

\begin{figure}
\caption{\label{fig:word_order_permutation_ring} The possible orders of S, O and V (blue) and their frequencies (red) according to \citet{Hammarstroem2016a}. Frequencies are measured in languages (left) and in families (right). }
\includegraphics[width = 0.5\textwidth]{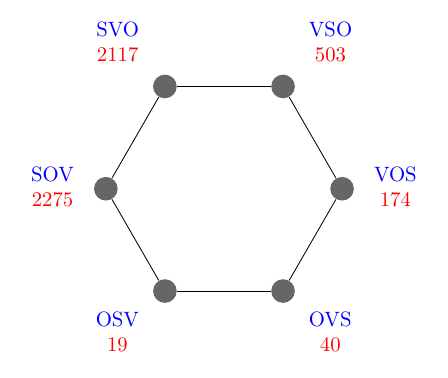}
\includegraphics[width = 0.5\textwidth]{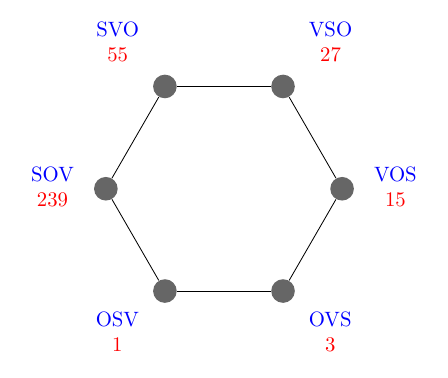}
\end{figure}

In this article, we have focused on predicting the optimal placement of the noun in the noun phrase. Do the arguments generalize to other heads?  
If we instead consider the optimal placement of V in the triple formed by S, O and V, the principle of surprisal minimization predicts that V should be placed at one of the ends. 
Let us consider the frequency of the dominant orders of the six possible orders from Hammarström's summary \citep{Hammarstroem2016a} and reproduced in Fig. \ref{fig:word_order_permutation_ring}. It turns out that the frequency of verb final or verb initial orders is significantly large when measured in languages ($58\%$ of languages) and also when it is measured in families ($83\%$ of languages). \footnote{
The one-sided binomial test described in Section \ref{subsec:preference_for_head_initial_or_final} with parameters $F$ and $p_{1,n} = 2/3$ for $n = 3$, yields a $\pvalue$ of $2.7\cdot 10^{-30}$ for counts in languages ($F = 5128$ and $g_{1,n}=2971$) while it yields a $\pvalue$ of $9.2\cdot 10^{-37}$ for counts in families ($F = 340$ and $g_{1,n}=282$).
These conclusions are obtained excluding languages/families lacking a dominant order from the counts. Integrating these languages/families properly in the binomial test, if possible, is beyond the scope of the present article. 
}

However, surprisal minimization alone cannot explain why the two most frequent orders are SOV and SVO by far \citep{wals-81,Hammarstroem2016a}. SOV is expected to be very frequent as it places the verb at the end as expected by surprisal minimization but OSV is the least frequent dominant order, apparently contradicting that principle. In addition, the placement of V at the end is much more frequent than the placement of the verb at the beginning 
(Fig. \ref{fig:word_order_permutation_ring})
but this can be fixed by postulating a preference for S first or by a strong preference to reduce the surprisal of V (rather than using V to reduce the surprisal of its arguments) because of its difficulty for learning and processing \citep{Ferrer2013f}. 
Similarly, syntactic dependency distance minimization alone neither can explain why the two most frequent orders are SOV and SVO by far. 
That principle predicts that V should be put at the center \citep{Ferrer2008e}; although SVO is the 2nd most attested word order, its mirror, OVS, has a very low frequency (Fig. \ref{fig:word_order_permutation_ring}), apparently contradicting syntactic dependency distance minimization. The problems above can be fixed with solutions of increasing number of additional principles required \footnote{What follows is just a synthesis of relevant components of the theory of word order developed by Ferrer-i-Cancho. Essential references are \citet{Ferrer2013e,Ferrer2013f,Ferrer2016c}}
\begin{enumerate}
\item
Invoking a third principle, i.e. swap distance minimization, that predicts that SVO and SOV form an economic partnership among all optimal orders predicted by surprisal minimization and minimization of the surprisal of the verb. 
The reason is that SVO and SOV are adjacent in the space of all permutations of subject, object and verb, that shows a ring structure (Figure \ref{fig:word_order_permutation_ring}). An edge joining two orders in the permutation ring indicates that one can be transformed into the other by just one swap of adjacent constituents (Figure \ref{fig:word_order_permutation_ring}).
The principle of swap distance minimization postulates that word order variants that require fewer swaps of adjacent elements to be produced from a source order are cognitively easier \citep{Ferrer2016c,Ferrer2023a,Franco2024a}.  Recall that the principle of syntactic dependency distance minimization and the principle of surprisal minimization are in conflict, namely their respective optimal orders are incompatible. Now suppose that a word order varies in such a way that it satisfies one of the two principles in conflict but it has also to be ``close'' to an order that satisfies the other order in conflict. 
Then the solution is that languages should be primarily SOV or SVO and if not, they should be close to them and these is what the frequencies of the six possible orders are suggesting (Figure \ref{fig:word_order_permutation_ring}), with the exception that OSV is close to SVO but it is the least frequent dominant order. 
\item
Invoking a third principle, i.e. a preference for S first, that would explain the high-frequency of SOV and SVO but this solution cannot explain why VSO and VOS have a frequency that cannot be neglected (Figure \ref{fig:word_order_permutation_ring}).
\item
Combining the two previous solutions, namely swap distance minimization and a subject (agent) first bias \citep{Sauppe2023a}, so as to have a more accurate account of the low number of OSV languages and the whole frequency distribution as a whole (Figure \ref{fig:word_order_permutation_ring}).
\end{enumerate} 
We have shown how swap distance minimization accounts for why the most attested dominant word orders fail to cover all optimal solutions predicted by syntactic dependency distance minimization and surprisal minimization of the verb. Swap distance minimization appears to be crucial but, have we simply fixed the problem by overfitting the data? We present two parallel lines of observations on how swap distance minimization illuminates word order variation synchronically and diachronically. Synchronically, 
\begin{itemize}
\item
When languages exhibit a couple of primary alternating orders of subject-object-verb, instead of one, the most frequent pairings involve orders that are adjacent in the word permutation ring (in order of decreasing frequency): SOV-SVO, VSO-VOS, SVO-VSO,...\citep{Ferrer2016c}. Notice that the second most attested pairing, VSO-VOS, goes against the subject first bias.   
\item
SOV is a cognitively primary or natural order \citep{Goldin-Meadow2008a}. In our framework this can be justified as SOV resulting from minimization of the surprisal of the verb and a subject first bias in conditions where syntactic dependency distance minimization can be neglected \citep{Ferrer2014a,Ferrer2013f}. With respect to previous research, we have refined these conditions in Section \ref{sec:introduction}. 
\item
The frequency of word orders tends to decrease as one moves in a clockwise sense from SOV in the permutation ring (Figure \ref{fig:word_order_permutation_ring}). 
\end{itemize}
Diachronically, 
\begin{itemize}
\item
SOV has been argued to be an early stage in word order evolution \citep{Newmeyer2000,Gell-Mann2011a}. 
\item
The main path of word order evolution has been argued to be SOV $\rightarrow$ SVO $\rightarrow$ VSO/VOS \citep{Gell-Mann2011a}, which implies shifts to adjacent orders in the permutation ring, except SVO to VOS, that requires two steps (Figure \ref{fig:word_order_permutation_ring}).  
\end{itemize}
Finally, these observations strongly suggest that evolutionary constraints (e.g., starting from SOV and path dependence) and the shape of the permutation ring may be also limiting the access to certain optimal orders predicted by syntactic dependency distance minimization and surprisal minimization of the verb.

In sum, the principle that we have applied for the optimal placement of the noun also generalizes to the case of the order of S, O and V but it requires to be integrated into a broader theoretical apparatus in order to understand how a few word order principles produce the distribution of dominant word orders.

\subsection{Sufficient versus necessary condition}

Here we have examined the prediction of a head initial or head final order when two conditions are met, namely (a) the sequence is short and (b) words are short, in the context of single head syntactic dependency structures.
In this formulation, the combination of these two conditions has the role of sufficient condition for a head initial/final order. We do not mean that the converse, i.e. a head initial/final order implies these two conditions hold. Put differently, we do not mean these two conditions are a necessary condition for a head initial/final order. To demonstrate that the converse is unlikely to be true, the ideal setting would be the dominant order of the quadruplet formed by a subject, an object, verb and an indirect object or oblique (SOVI or SOVX) in a diverse sample of languages. 
In this setting, (a) the sequences would have $n \geq 4$ because each of these four constituents may be formed by more than one word and thus the sentence length will never be shorter than in Dryer's noun phrase setting and (b) the words are likely to be longer because the head of each constituent is likely a content word (a noun in the S, O and I/X constituent, and a verb in the V constituent); at least four words of the sequence are likely to be long compared to the demonstrative and the high frequency numerals that will typically form the noun phrase that we have investigated here.
As these ideal data are not readily available,
we approximate it via the dominant order of just S, O and V, with the sole purpose of providing a quick counterexample for the converse. In this setting, (a) the sequence may be longer than $n=4$ in general because each of the three constituents may contribute with more than one word and (b) the words are likely to be longer because the head of each constituent is likely a content word. In Section \ref{subsec:generalization}, we have already shown preliminary evidence of a preference to place the verb first or last in triples formed by  S, O and V.

\subsection{Future work}

In this article we have tested the predictions of the optimal placement of the nominal head based on frequencies of preferred orders but a gradient approach would be more powerful \citep{Levshina2023a}. Future research should extend the present analyses to other single head structures but also to frequencies that correspond to corpus frequencies as in previous research \citep{Ferrer2019a,Ferrer2020b} and distinguishing between star trees rooted at the hub vertex (namely the single head structures we have studied in this article) from star trees rooted on a leaf vertex as \citet{Courtin2019a} did.   

In this article we have investigated the optimal placement of the head in a reductionist fashion, i.e. independently of other parts of the syntactic dependency structure and also independently of evolution. To illustrate this, consider the case of S, O and V structures. Concerning other parts of the sentence, the principle of syntactic dependency minimization predicts the optimal placement of nominal heads as a function of the placement of the verb: in verb final orders, it predicts noun first placement in S and O (left branching); in verb initial orders, it predicts noun last placement in S and O (right branching) \citep{Ferrer2013e}. Concerning evolution, it has been hypothesized that SOV languages should tend to put nouns first to prevent regression to SOV \citep{Ferrer2013e}. Future research should explore the interaction between the placement of nominal heads and the order of higher level constituents such as S, O and V. 

\iftoggle{anonymous}{}{

\section*{Acknowledgments}

We are very grateful to L. Alemany-Puig for a careful revision the manuscript and helpful comments. This article has benefited from the very valuable feedback of two anonymous reviewers. A brief summary of the article was presented during the Scientific Dialogue on \href{https://www.ae-info.org/ae/Acad_Main/Events/The\%20Bases\%20of\%20Language}{\em ``The bases of language: from tabula rasa to biological determinism''} organized by Academia Europaea (Barcelona Knowledge Hub) on 22 November 2023. The present work was also presented in the Autogramm seminar at Sorbonne Nouvelle (1 February 2024, Paris) and the COLT seminar at Pompeu Fabra University (21 February 2024, Barcelona). The author is grateful to the organizers and participants of these events for valuable questions and discussions. This research is supported by a recognition 2021SGR-Cat (01266 LQMC) from AGAUR (Generalitat de Catalunya) and the grants AGRUPS-2022 and AGRUPS-2023 from Universitat Politècnica de Catalunya.
}

\bibliographystyle{apacite}

\bibliography{../../../../Dropbox/biblio/rferrericancho,../../../../Dropbox/biblio/complex,../../../../Dropbox/biblio/ling,../../../../Dropbox/biblio/cl,../../../../Dropbox/biblio/cs,../../../../Dropbox/biblio/maths}

\end{document}